\pdfoutput=1

\documentclass[11pt]{article}

\usepackage[review]{acl}

\usepackage{times}
\usepackage{latexsym}

\usepackage[T1]{fontenc}

\usepackage[utf8]{inputenc}

\usepackage{microtype}

\usepackage{inconsolata}

\usepackage{graphicx}
\usepackage{tabularx}
\usepackage{array}
\usepackage{adjustbox}
\usepackage{multirow}
\usepackage{longtable}
\usepackage{colortbl}

\usepackage{enumitem,pifont,xcolor}
\usepackage{graphicx}
\usepackage{amsmath}
\usepackage{booktabs}
\usepackage{svg}

\title{First Multi-Dimensional Evaluation of Flowchart Comprehension \\ for Multimodal Large Language Models}

\author{First Author \\
  Affiliation / Address line 1 \\
  Affiliation / Address line 2 \\
  Affiliation / Address line 3 \\
  \texttt{email@domain} \\\And
  Second Author \\
  Affiliation / Address line 1 \\
  Affiliation / Address line 2 \\
  Affiliation / Address line 3 \\
  \texttt{email@domain} \\}

\begin{document}
\maketitle
\begin{abstract}
With the development of Multimodal Large Language Models (MLLMs) technology, its general capabilities are increasingly powerful. To evaluate the various abilities of MLLMs, numerous evaluation systems have emerged. But now there is still a lack of a comprehensive method to evaluate MLLMs in the tasks related to flowcharts, which are very important in daily life and work. We propose the first comprehensive method, FlowCE, to assess MLLMs across various dimensions for tasks related to flowcharts. It encompasses evaluating MLLMs' abilities in Reasoning, Localization Recognition, Information Extraction, Logical Verification, and Summarization on flowcharts. However, we find that even the GPT4o model achieves only a score of 56.63. Among open-source models, Phi-3-Vision obtained the highest score of 49.97. We hope that FlowCE can contribute to future research on MLLMs for tasks based on flowcharts.
\end{abstract}

\section{Introduction}

In the modern work environment, flowcharts have become a widely used graphical tool across various industries and fields. Flowcharts provide an intuitive and efficient way to describe and analyze workflows. By representing processes graphically, complex workflows can be simplified into easily understandable steps, thereby facilitating a range of tasks. Currently, leveraging Multimodal Large Language Models (MLLMs) for the understanding and analysis of flowcharts has become a research focus. Represented by models like GPT-4v \cite{gpt4}, these large models can comprehend user-input images and perform corresponding question-and-answer tasks. Meanwhile, there have been numerous open-source efforts for MLLMs, such as LLAVA-1.6v \cite{lava1.6}, QWEN-VL \cite{qwenvl}, MiniCPM \cite{minicpm}, phi-3-vision \cite{phi3}, and CogVLM2 \cite{cogvlm}. To evaluate the cross-modal understanding capabilities of existing MLLMs between images and text, various benchmarks have emerged, including MMBench \cite{mmbench}, MME \cite{mme}, TextVQA \cite{singh2019towards}, MM-Vet \cite{mm-vet}, DocVQA \cite{docvqa}, ChartQA \cite{chartqa}, InfographicQA \cite{infographicvqa}, FlowChartQA \cite{flowchartqa} and so on. Additionally, these evaluation systems measure the capabilities of MLLMs from different perspectives, including the understanding of general images, document-type images, chart-type images, and more.

\begin{figure}[t]
\hspace*{-6.0mm}
\centering
\includegraphics[width=3.2in]{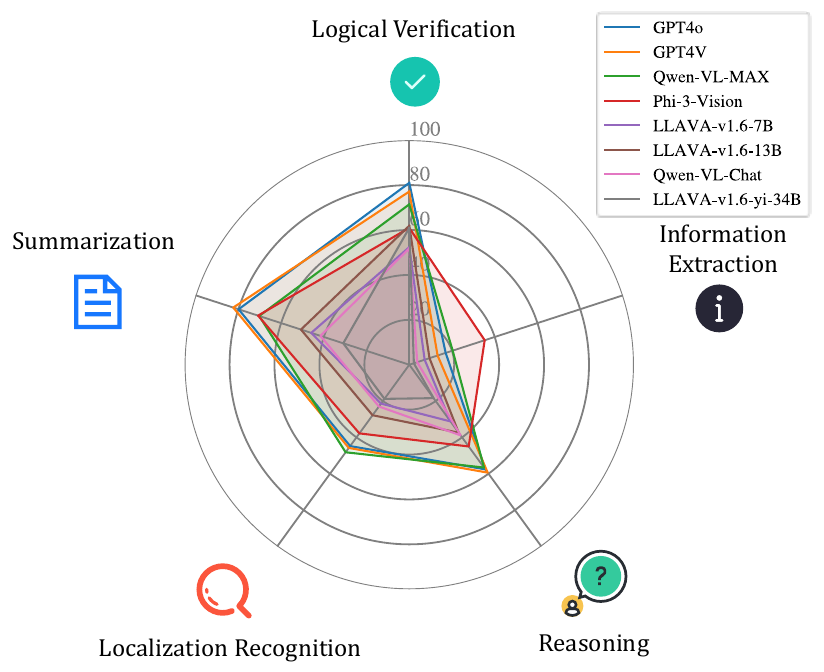}
    \caption{Evaluation results of multimodal large language models on five dimensions of tasks in FlowCE. GPT-4o achieves the highest overall score of 56.63.}
\label{fig:1}
\end{figure}

\begin{table*}[!ht]
    \centering
    \small
    \adjustbox{max width=\textwidth}{
    \begin{tabular}{ccccccccccc}
    \hline
        Benchmark & Capability & Real-world Data & Handcrafted Questions & Answer Type & Size & \# models  \\ \hline
        LVLM-eHub & General Multi-Modality & \ding{51} &  \ding{115} & MC/OE & 332k &  8 \\ 
        MME & General Multi-Modality & \ding{51} & \ding{51} & MC & 2,194 &  10  \\ 
        MMBench & General Multi-Modality & \ding{51} &  \ding{115} & MC & 2,974 &  14  \\ 
        TextVQA & Text Recognition and  Contextual Reasoning & \ding{51} & \ding{51} & OE & 45.3k &  6 \\ 
        InfographicVQA & Integrated Document Visual and Textual Reasoning & \ding{51} & \ding{51} & OE & 30k &  1  \\ 
        ChartQA & Chart Understanding and Analysis & \ding{51} & \ding{51} & OE & 9.6k &    4 \\ 
        EgoThink & First-Person Thinking & \ding{51} & \ding{51} & OE & 700 &  21  \\ 
        MathVista & Mathematical Reasoning & \ding{51} & \ding{51} & MC & 6141 &  11  \\ 
        FlowchartQA & Geometirc and Topological Information of Flowcharts & \ding{55} & \ding{55} & MC & 6M &   1 \\ 
        FlowCE(ours) & Comprehensive Understanding of Flowcharts   & \ding{51} & \ding{51} & OE & 505 & 19 \\ \hline
    \end{tabular}
    }
    \caption{Comparison of recent comprehensive evaluation benchmarks of MLLMs and our proposed benchmark FlowCE. MC/OE indicate multi-choice and open-ended question-answering respectively. "\ding{115}" indicates that there are both handcrafted questions and questions generated using templates.  }
    \label{table:detailed-dataset-info}
\end{table*}

However, to the best of our knowledge, none of these existing evaluation benchmarks comprehensively assess MLLMs' understanding of flowcharts from multiple perspectives in real-world scenarios. This hinders the development of methods for utilizing MLLMs to understand and analyze flowcharts in open environments. Thus, inspired by previous works such as FigureQA \cite{figureqa}, PlotQA \cite{plotqa}, ChartQA \cite{chartqa} and FlowchartQA \cite{flowchartqa}, and motivated by the successful development of MLLMs. We propose a novel benchmark, FlowCE, for the first time: comprehensively assessing the understanding capabilities of multimodal large language models on flowcharts in real-world scenarios. FlowCE evaluates the understanding capabilities of existing MLLMs on flowcharts from multiple dimensions, including Reasoning, Information Extraction, Localization Recognition, Summarization, and 
Logical Verification. We have carefully designed diverse question-answer pairs for various dimensional tasks in open environments. Additionally, the flowchart images in FlowCE are sourced from a variety of real-world scenarios and styles. We have carefully designed diverse question-answer pairs for various dimensional tasks in open environments. Additionally, the flowchart images in FlowCE are sourced from a variety of real-world scenarios and styles. 

We conducted evaluations on all mainstream MLLMs, both open-source and proprietary, using FlowCE. The evaluation results for some parts on FlowCE are shown in Figure \ref{fig:1}. The results indicate that even the highly performant GPT4o achieves only a score of 56.63, with the best performance among open-source models being achieved by Phi-3-Vision \cite{phi3}, scoring 49.97. Our main contributions are as follows:

\begin{itemize}
    {
    \item 
    We introduce FlowCE to comprehensively evaluate the understanding capabilities of MLLMs on flowcharts. It encompasses evaluation tasks and methodologies across dimensions such as reasoning, information extraction, localization recognition, summarization, and logical verification.
    \item 
    We extensively evaluate mainstream open-source and proprietary models using FlowCE. Through detailed analysis of these MLLMs' performance across different dimensional tasks, we uncovered their strengths and limitations in understanding flowcharts. Additionally, we proposed some improvement suggestions for existing models to facilitate future research and development. 
    \item 
    We are open-sourcing our resources to foster future advancements in this field.
    }
\end{itemize}

\section{Related Work}

\subsection{Multimodal Large Language Models}

Inspired by the remarkable success of LLMs such as internVL \cite{chen2024internvl}, llama3 \cite{touvron2023llama}, Yi-chat \cite{young2024yi}, Qwen \cite{qwen}, and Vicuna \cite{zheng2024judging_vicuna}, recent MLLMs have incorporated these advanced LLMs as their primary backbone. Examples include the LLAVA-V1.6 \cite{liu2024llava} series, ShareGPT4 \cite{chen2023sharegpt4v} series, Qwen \cite{qwenvl} series, Cogvlm \cite{cogvlm} series and so on. Initially, MLLMs leverage vast datasets consisting of image-text pairs \cite{alayrac2022flamingo,zhu2024multimodal} or an arbitrarily combination of visual and textual data for pre-training \cite{li2023mimic, liu2024improved}. Moreover, the availability of extensive image-text instruction datasets facilitate recent studies \cite{dai2024instructblip, li2023mimic,liu2024visual,ye2023mplug,chen2023sharegpt4v} to implement instruction tuning. This fine-tuning process enhances the ability of MLLMs to produce high-quality responses. This two-phase training strategy  \cite{li2022overcoming, yang2022multimodal} enables recent MLLMs to achieve outstanding performance in downstream vision-language tasks \cite{antol2015vqa, hudson2019gqa, lin2014microsoft, plummer2015flickr30k}.

\subsection{Benchmarks for MLLMs}

To evaluate the capabilities of Vision-Language Models (MLLMs), various downstream language tasks are employed. General benchmarks, such as MMBench \cite{mmbench}, MME \cite{mme}, and LVLM-ehub \cite{Lvlmehub}, provide a comprehensive assessment of model performance. Domain-specific benchmarks, such as TextVQA \cite{singh2019towards} and DocVQA \cite{docvqa}, evaluate the fine-grained transcription capabilities of MLLMs on low-resolution images. MathVista \cite{lu2024mathvista} examines the ability of MLLMs to integrate visual and mathematical logic. ChartQA \cite{chartqa} aims to evaluate direct chart understanding and analysis, while InfographicQA \cite{infographicvqa} addresses logical questions about data visualizations and charts. EgoThink \cite{cheng2024egothink} elaborate on the capabilities of MLLMs to think from a first-person perspective. General benchmarks offer a broad and consistent evaluation framework \cite{Lvlmehub, mme, mmbench}, whereas domain-specific benchmarks enable detailed assessment of model capabilities and promote advancements in specific research areas.

In Table \ref{table:detailed-dataset-info}, we compare FlowCE with various existing benchmarks. FlowCE comprehensively assesses for the first time the ability of MLLMs to understand flowcharts. Specifically, compared to FlowchartQA \cite{flowchartqa}, we not only introduce tasks across more dimensions but also create real-world flowchart data and open-scenario question-answer pairs.

\begin{figure*}[htb]
    \centering
    \includegraphics[width=\textwidth]{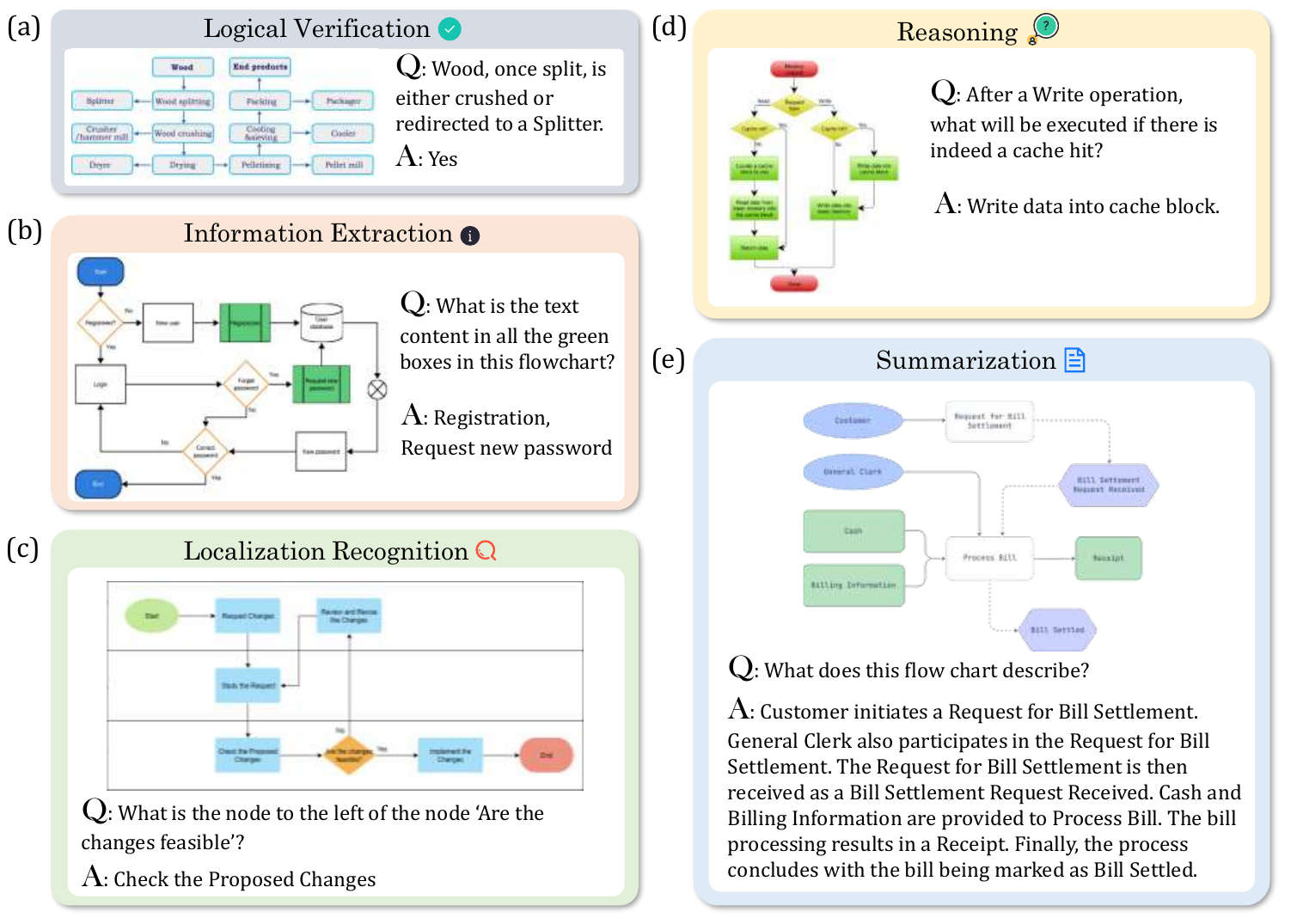} 
    \caption{Data samples of FlowCE, which covers 5 evaluation dimensions. Each evaluation dimension contains human-annotated question-answer pairs.}
    \label{fig:FlowCE-example}
\end{figure*}

\section{FlowCE}
In this section, we elaborate first on the evaluation tasks across various dimensions in FlowCE. Then, we introduce the process of manually constructing diverse open-scenario question-answer pairs. Finally, we present the evaluation methodologies for tasks across different dimensions.

\subsection{Tasks across different dimensions}
As shown in Figure \ref{fig:FlowCE-example}, we establish tasks across five dimensions in real flowchart scenarios, including reasoning, information extraction, localization recognition, summarization, and logical verification, for quantitative evaluation.
\\
\textbf{Logical Verification} \quad  Upon receiving a process diagram, users provide the logical relationships between different nodes or boxes in the diagram, and MLLMs are tasked with evaluating these relationships. Figure \ref{fig:FlowCE-example}(a) shows an example of Logical Verification. This process entails a comprehensive analysis of the structure and content of the process diagram to ensure the accuracy and coherence of the logical relationships. MLLMs assess whether the provided process logic aligns with the actual scenario by considering the interactions among individual nodes and their roles throughout the entire process.
\\
\textbf{Information Extraction} \quad The task entails MLLMs receiving flowchart images and extracting corresponding textual information based on user queries. We have categorized the questions into two main types based on the content of the flowchart: the first type involves prompting MLLMs to extract all textual information from the flowchart, while the second type entails extracting specific textual information based on the characteristics of the flowchart. An example of Information Extraction is shown in Figure \ref{fig:FlowCE-example}(b).
\\
\textbf{Localization Recognition} \quad  Users will inquire about the positional relationships between different nodes or boxes in the flowchart, an example of Localization Recognition is illustrated in Figure \ref{fig:FlowCE-example}(c), thereby assessing whether MLLMs have an accurate understanding of the positional relationships of nodes and boxes in the flowchart.
\\
\textbf{Reasoning} \quad For an example of Reasoning, as shown in Figure \ref{fig:FlowCE-example}(d), the task refers to MLLMs making decisions in response to user inquiries based on the content of the flowchart images after being provided with them. Here, we formulate more natural questions based on the content of the flowchart, which require judgment and reasoning considering aspects such as conditional relationships within the flowchart to answer, rather than relying solely on the direction of the arrows in the flowchart.
\\
\textbf{Summarization} \quad  MLLMs provide a summarized abstraction of the content depicted in process diagrams, elucidating the conveyed information. They accomplish this task by analyzing the logical relationships among various nodes within the diagram, identifying key steps and critical information, and integrating them into a concise yet comprehensive summary. Through understanding and encapsulating the process diagram, MLLMs generate the primary flow of the process and key decision points, thereby assisting users in better comprehending the process or system represented by the diagram, as shown in Figure \ref{fig:FlowCE-example}(e).

\subsection{Data construction}
\begin{figure}
    \centering
    \includegraphics[width=1.0\linewidth]{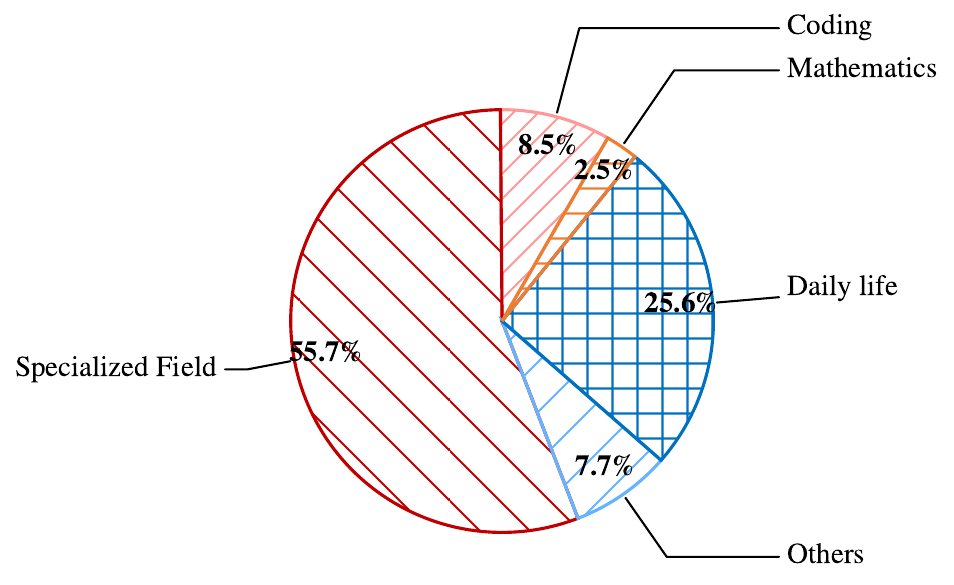}
    \caption{Distribution of Different Types of Flowcharts}
    \label{distribution}
\end{figure}

In this section, we introduce the data of FlowCE and elaborate on the detailed process of constructing FlowCE.
\\
\textbf{FlowCE-data} \quad FlowCE is built upon 500 real-world flowcharts, ensuring an ample diversity in each chart. In Figure \ref{distribution}, we present a detailed breakdown of the category distribution within the flowchart, encompassing categories from daily life, various specialized filed flowcharts, coding flowcharts, mathematical flowcharts, and others.
\\
\textbf{Human-annotated} \quad To ensure an open-ended question-and-answer format, we manually constructed question-answer pairs for each flowchart. We assigned different dimensions of tasks to the same individual to annotate a particular type of question, ensuring consistency in the tasks. Additionally, to allow for greater diversity in open-ended question-and-answer scenarios, we leveraged powerful GPT-like models for auxiliary construction, aiding humans in exploring more imaginative possibilities. Please refer to the Appendix \ref{sec:appendix} for specific details.

\renewcommand{\arraystretch}{1.4} 
\begin{table*}[!ht]
    \centering
    \large
     \adjustbox{max width=\textwidth}{
    \begin{tabular}{ccccccc}
    \hline
        \multirow{2}{*}{\textbf{Models}} & \multirow{2}{*}{\textbf{Image Encoder}} & \multirow{2}{*}{\textbf{LLM}} & \multirow{2}{*}{\textbf{Alignment Module}} & \multirow{2}{*}{\textbf{ToP}} & \multicolumn{2}{c}{\textbf{Dataset Size}} \\ \cline{6-7}
    & & & & & Pretraining & Finetuning \\ 
        \hline
        \rowcolor{gray!5}
        \multicolumn{7}{c}{\textbf{API-based (Proprietary) Models}}  \\
        \hline

         GPT4o \cite{gpt4} & \multicolumn{6}{c}{\multirow{3}{*}{/}} \\ 
         GPT4V \cite{gpt4} & \multicolumn{6}{c}{} \\ 
         Qwen-VL-MAX  \cite{qwenvl} & \multicolumn{6}{c}{} \\ 

        \hline
        \rowcolor{gray!5}
        \multicolumn{7}{c}{\textbf{3.4B\textasciitilde 7B Models}}  \\
        \hline
        MiniCPM-V2 \cite{minicpm} & SigLip-400M & MiniCPM-2.4B & RLHF-V \cite{yu2024rlhf} & 3.43B & / & / \\ 
        Phi-3-Vision \cite{phi3} & CLIP-ViT-L-336px & phi-3-mini-128K-instruct & SFT+DPO & 4.2B & 0.5T & 15B \\ 
        LLAVA-V1.5-7B \cite{liu2024improved} & CLIP-ViT-L-336px & Llama2-7B & MLP & 7.1B & 558K & 665K \\ 
        ShareGPT4V-7B \cite{chen2023sharegpt4v} & CLIP-ViT-L-336px & Vicuna-7B & MLP & 6.7B & 1.2M & 665K \\ 
        LLAVA-V1.6-7B \cite{liu2024llava} & CLIP-ViT-L-336px & Vicuna-7B & Linear & 7.06B & 558K & 760K \\ 
        \hline
        \rowcolor{gray!5}
        \multicolumn{7}{c}{\textbf{8B\textasciitilde 13B Models}}  \\
        \hline
        LLAVA-Llama3-8B \cite{contributors2023xtuner}  & CLIP-ViT-L-336px & Llama3-8B-Instruct & MLP & 8.03B & 558K & 665K \\ 
        MiniCPM-Llama3-V2.5 \cite{minicpm} & SigLip-400M & Llama3-8B-Instruct & RLAIF-V \cite{yu2024rlaif}& 8.54B & / & / \\ 
        Qwen-Chat-VL \cite{qwenvl} & Open-CLIP-bigG & Qwen-7B & Cross-Attention & 9.6B & 1.4T & 76.8M \\ 
        LLAVA-V1.5-13B \cite{liu2024improved} & CLIP-ViT-L-336px & Llama2-13B & MLP & 13.3B & 558K & 665K \\ 
        ShareGPT4V-13B \cite{chen2023sharegpt4v} & CLIP-ViT-L-336px & Vicuna-13B & MLP & 12.58B & 1.2M & 665K \\ 
        LLAVA-V1.6-13B \cite{liu2024llava} & CLIP-ViT-L-336px & Vicuna-13B & Linear & 13.3B & 558K & 760K \\ 
        \hline
        \rowcolor{gray!5}
        \multicolumn{7}{c}{\textbf{13B\textasciitilde  Models}}  \\
        \hline
        Cogvlm-Chat \cite{cogvlm} & EVA2-CLIP-E & CogVLM-17B & Visual Expert & 17.6B & 1.5B & / \\ 
        Cogvlm2-Llama3-Chat-19B \cite{cogvlm} & EVA2-CLIP-E & Meta-Llama-3-8B-Instruct & Visual Expert & 19.5B & / & / \\ 
        LLAVA-Internlm2-Chat-20B \cite{contributors2023xtuner} & CLIP-ViT-L-336px & InternLM2-Chat-20B & deepspeed finetuning & 20B & 595K & 150K \\ 
        LLAVA-Next-Yi-34B \cite{liu2024llava} & CLIP-ViT-L-336px & Nous-Hermes-2-Yi-34B & Linear & 34.8B & 558K & 760K \\ 
        Yi-VL-34B \cite{young2024yi} & CLIP-ViT-L-336px & Yi-34B-Chat & MLP & 34B & 3.1T & 1.25M \\ \hline
    \end{tabular}
        }
        \caption{Statistics of compared API-based and open-source MLLMs, where ToP indicates Total Parameters and '/' indicates no detailed information for now.}
        \label{table:model-training-dataset}
\end{table*}

\subsection{Evaluation method}
In this section, the evaluation of various tasks quantification methods will be introduced.
\\
\textbf{Automatic evaluation} \quad For tasks involving open-ended question answering, such as reasoning, localization recognition, and summarization, we employ GPT4 to assess the semantic similarity between standard answers and the responses generated by MLLMs. For detailed methodology of the evaluation utilizing GPT4, please refer to the Appendix \ref{appendix:Large Language Model Judge}.
\\
\textbf{Accuracy calculation} \quad Firstly, for the logical verification task, we match the output of MLLMs, either \textit{"Yes"} or \textit{"No,"} with the standard answers to calculate the accuracy after all questions have been answered, thereby quantifying the score of MLLMs on this task. Next, for the information extraction task, we propose a method based on effective factor to fairly compare the content generation effectiveness of different MLLMs. Then, for the information extraction task, we propose a method based on the effective factor to fairly compare the performance of different MLLMs in generating content. Suppose the label set is given by $\text{label} = [\text{text}_1, \text{text}_2, \ldots, \text{text}_n]$, where $\text{text}_n$ represents the $n$-th text. The output answers are given by $\text{prediction} = [\text{pre}_1, \text{pre}_2, \ldots, \text{pre}_m]$, where $\text{pre}_m$ is the $m$-th predicted text. If there is a predicted text in $\text{prediction}$ that does not exist in $\text{label}$, and there are $t$ such texts ($t \geq 1$), then the effective factor $\delta$ changes according to the following formula:
\[
\delta = \delta^t,
\]
At this point, if there is a predicted text in $\text{prediction}$ that exists in $\text{label}$, then the initial score $s$ changes as follows:
\[
s = s \cdot \delta,
\]

If $t = 0$, then for each predicted text in $\text{prediction}$ that exists in $\text{label}$, the score remains the initial score $s$. Suppose there are $i$ predicted texts that exist in $\text{label}$, the total score is $s \cdot i$. The product of the number of texts in $\text{label}$ and the initial score is denoted as $a$. The quantitative score for evaluating MLLMs on this task is given by:
\[
\text{Score} = \frac{s \cdot i}{a} (\%).
\]

\section{Experiments}

\begin{table*}[h]
\centering
\renewcommand{\arraystretch}{1.2}
\setlength{\tabcolsep}{3pt}
\begin{adjustbox}{max width=\textwidth}
\small
\begin{tabularx}
{\textwidth}
{>{\centering\arraybackslash}m{4.5cm}>{\centering\arraybackslash}X>{\centering\arraybackslash}X>{\centering\arraybackslash}X>{\centering\arraybackslash}X>{\centering\arraybackslash}X>{\centering\arraybackslash}m{1.25cm}}
\hline
\multirow{2}{*}{\textbf{Models}} & \multicolumn{5}{c}{\textbf{FlowCE}} & \multirow{2}{*}{\textbf{Average}} \\
\cline{2-6}
& \textbf{LV} & \textbf{IE} & \textbf{RS} & \textbf{LR} & \textbf{SM} & \\
\hline
\rowcolor{gray!5}
\multicolumn{7}{c}{\textbf{API-based Models}}  \\
\hline
GPT4o & \textbf{83.81} & \underline{17.04} & \underline{57.60} & 44.80 & \underline{79.90} & \color{red}{56.63} \\
GPT4V & \underline{77.14} & 12.94 & \textbf{59.40} & \underline{45.80} & \textbf{82.40} & 55.54 \\
Qwen-VL-MAX & 72.38 & \textbf{20.32} & 56.60 & \textbf{48.20} & 70.25 & 53.55 \\

\hline
\rowcolor{gray!5}
\multicolumn{7}{c}{\textbf{3.4B\textasciitilde 7B Models}}  \\
\hline
MiniCPM-V2 & 51.43 & 7.00 & 30.00 & \underline{22.00} & \underline{50.20} & 32.13 \\
Phi-3-Vision  & \textbf{60.95} & \textbf{35.30} & \textbf{45.00} & \textbf{37.80} & \textbf{70.80} & \color{blue}{49.97} \\

LLAVA-V1.5-7B & \underline{53.33} & 4.90 & 14.40 & 18.20 & 35.60 & 25.29 \\
ShareGPT4V-7B & 50.48 & 3.72 & 12.20 & 16.80 & 33.60 & 23.36 \\
LLAVA-V1.6-7B  & 52.38 & \underline{7.20} & \underline{31.20} & 21.40 & 45.90 & 31.62 \\
\hline
\rowcolor{gray!5}
\multicolumn{7}{c}{\textbf{8B\textasciitilde 13B Models}}  \\
\hline
LLAVA-Llama3-8B & 55.24 & 8.04 & 21.20 & 20.80 & 33.20 & 27.70 \\
MiniCPM-Llama3-V2.5 & \underline{58.10} & \textbf{12.25} & \textbf{45.20} & \textbf{42.80 }& 17.20 & 35.11 \\
Qwen-Chat-VL & 50.48 & 3.73 & \underline{38.80} & 23.00 & \underline{41.60} & 31.52 \\
LLAVA-V1.5-13B & 53.33 & 5.36 & 22.60 & 22.20 & 40.50 & 28.80 \\
ShareGPT4V-13B & 53.33 & 4.46 & 22.20 & 16.60 & 41.50 & 27.62 \\
LLAVA-V1.6-13B  & \textbf{62.86} & \underline{9.47} & 37.40 & \underline{27.80} & \textbf{50.70} & 37.65 \\
\hline
\rowcolor{gray!5}
\multicolumn{7}{c}{\textbf{13B\textasciitilde Models}}  \\
\hline
Cogvlm-Chat & 50.48 & 0.34 & 34.80 & 29.60 & 53.20 & 33.68 \\
Cogvlm2-Llama3-Chat-19B & 57.14 & 4.70 & \underline{44.60} & \textbf{37.20} & \textbf{74.30} & 43.59 \\
LLAVA-Internlm2-Chat-20B & \underline{59.05} & \underline{5.69} & 15.40 & 19.00 & 41.90 & 28.21 \\
LLAVA-Next-Yi-34B & \textbf{60.95} & \textbf{12.21} & \textbf{51.20} & \underline{34.20} & \underline{63.10} & 44.33 \\
Yi-VL-34B & \textbf{60.95} & 2.14 & 18.40 & 18.80 & 30.90 & 26.24 \\
\hline
\end{tabularx}
\end{adjustbox}
\caption{Detailed evaluation results on FlowCE across different models, where "LV" stands for Logical Verification, "IE" for Information Extraction, "RS" for Reasoning, "LR" for Localization Recognition, and "SM" for Summarization. \textbf{Bold font} indicates the best performance in the same category, while \underline{underlined font} indicates the second-best performance in the same category. \color{red}{Red} \color{black}{ indicates the  highest average score among all API-based models.} \color{blue}{Blue} \color{black}{indicates the highest average score among all open-source models.}
}
\label{tab:performance}

\end{table*}

\begin{figure*}[htb]
    \centering
    \includegraphics[width=\textwidth]{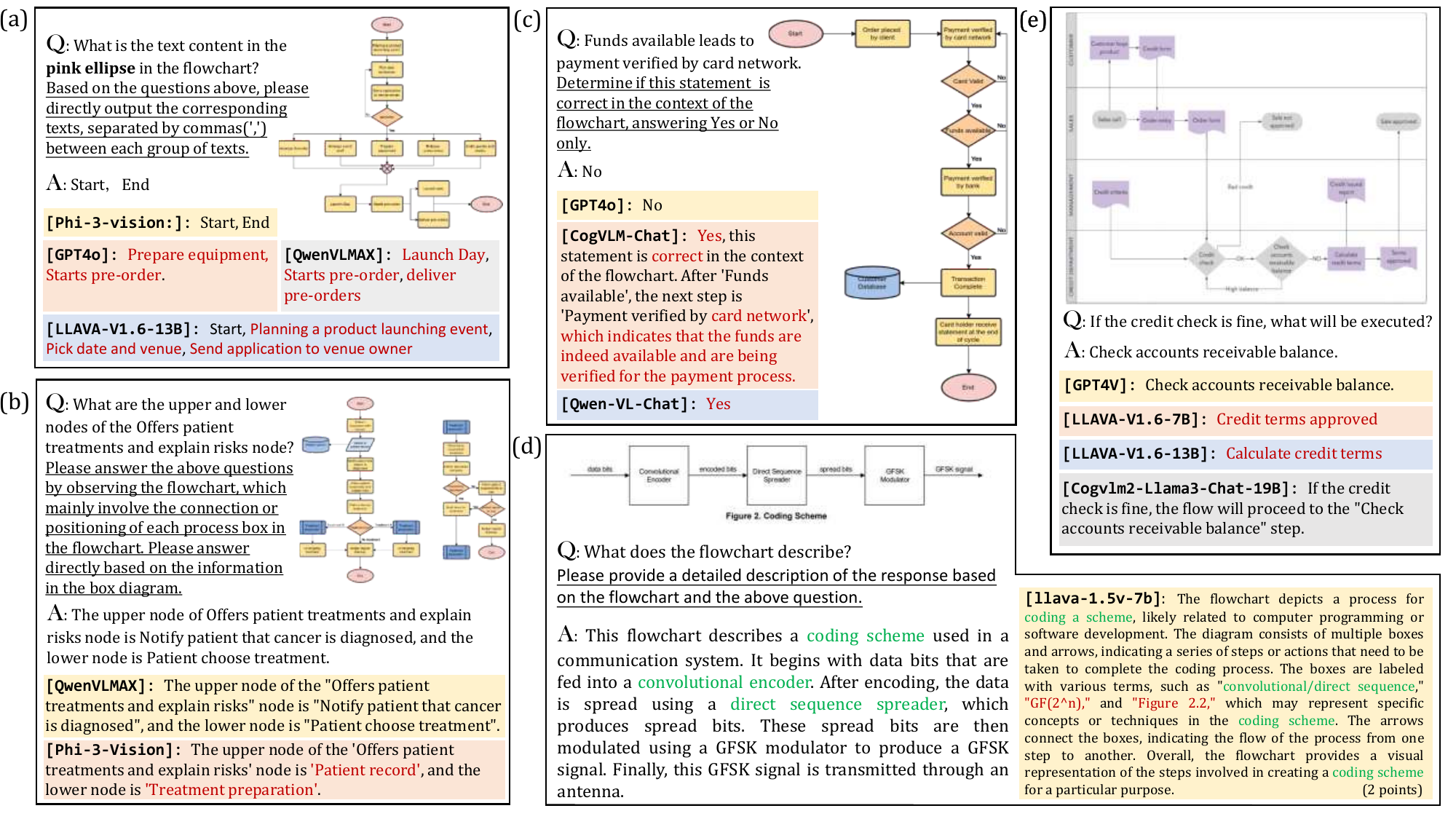} 
    \caption{Some results from vary MLLMs. \underline{The words underlined} indicate additional prompts. (a) showcases results on Information Extraction, (b) presents results on Localization Recognition, (c) showcases results on Logical Verification, (d) showcase results on Summarization, (e) displays results on Reasoning. }
    \label{model prediction example}
\end{figure*}

\subsection{Experimental setups}
We conduct experiments on existing mainstream MLLMs, including both proprietary and open-source models. The parameter sizes of the open-source models range from 3.4B to 7B, 8B to 13B, and above 13B. In Table \ref{table:model-training-dataset}, we provide a detailed overview of these evaluated models in our experiments.

We employ GPT-4 as the adjudicator for LLMs to assign evaluation scores, with a focus on semantic similarity between standard answers and MLLM model outputs. Our evaluations adhere to a protocol: for reasoning and localization recognition tasks, we set the score range per question from 0 to 5. For summarization tasks, the score range per question is from 1 to 10. In the evaluation of information extraction tasks, we set the score \( s \) as 2, with an effective factor \( \delta \) of 0.8.

\subsection{Evaluation results}

We extensively evaluate open-source MLLMs models at different parameter levels and mainstream commercial MLLMs models. All detailed evaluation results are presented in Table \ref{tab:performance}. Despite significant advancements in MLLMs in recent years, they still struggle to demonstrate understanding of flowcharts, including GPT-4o. Across five different task dimensions, only the summarization task achieves relatively high scores, peaking at 82.40 points in closed-source models. However, this is only demonstrated in closed-source models; in open-source models, the highest score reaches only 74.30 points. The highest score attained in the information extraction task is only 35.30 points, while in the reasoning task, it reaches a maximum of 59.40 points. In the localization recognition task, the highest score is 48.20 points. Even under random guessing with a score of 50.00 points in the logic validation task, the highest score reaches only 83.81 points. Among all closed-source models, GPT4o demonstrates superior overall capabilities compared to other models, but only excels in the logic validation task. Among all open-source models, Phi-3-Vision achieves the highest scores, surpassing closed-source models in the information extraction task. We will further elaborate on detailed assessments across different task dimensions. Additional cases can be found in the Appendix \ref{appendix:additional example}.
\\
\textbf{Results of Information Extraction} \quad 
In this task, models generally obtain very low scores. The highest score of 35.30 is achieved by Phi-3-Vision, with the second-place model being the proprietary model Qwen-VL-MAX, but only scoring 20.32, indicating a significant gap. In Figure \ref{model prediction example}(a), for instance, by highlighting the inherent feature "pink ellipse" in the flowchart, MLLMs are enabled to extract corresponding information, with only Phi-3-Vision producing the correct answer. In Appendix \ref{appendix:Informaton-Extraction-task}, to demonstrate the performance variation of different models in Information Extraction tasks, we conduct visual analysis based on effective factors. For example, Phi-3-vision achieves an average effective factor score exceeding 0.6.
\\
\textbf{Results of Localization Recognition} \quad The evaluation results of various models in this task indicate poor performance overall, with the top performer Qwen-VL-MAX scoring only 48.20 points. In Figure \ref{model prediction example}(b), detailed examples of Qwen-VL-MAX and Phi-3-Vision are presented. The response of Qwen-VL-MAX correctly identifies the upper and lower nodes of the \textit{"Offers patient treatments and explain risks"} node as \textit{"Notify patient that cancer is diagnosed"} and \textit{"Patient choose treatment"} respectively. This indicates a clear understanding of the flowchart and the ability to accurately identify the relationships between different nodes. On the other hand, Phi-3-Vision incorrectly identifies the upper node as \textit{"Patient record"} and the lower node as \textit{"Treatment preparation."} This suggests that Phi-3-Vision struggled with accurately interpreting the connections between the nodes in the flowchart, leading to an incorrect answer.
\\
\textbf{Results of Logical Verification} \quad 
For this tasks, the open-source models LLAVA-V1.6-13B, Phi-3-Vision, LLAVA-Next-Yi-34b, and Yi-VL-34B have achieved the top two performances. Regarding the highest scoring model, GPT4o, as depicted in Figure \ref{model prediction example}(c), it exhibits concise and clear responses to questions with stronger instruction-following capabilities. Conversely, models such as CogVLM-Chat tend to generate more hallucinatory descriptions in their answers, leading to erroneous judgments. For instance, in the case of Qwen-Chat-VL, it outputs answers of the \textit{"Unknown"} type, indicating a deficiency in instruction-following capability. In Figure \ref{bar} of Appendix \ref{appendix:predicted_vs_actual}, we also analyze the predictive distributions of different models and visually compare them with the distribution of true labels. We find that the predictions of most models exhibit significant biases in this task. For example, ShareGPT4V-7B categorizes all results as correct. Only GPT4v, GPT4o, LLava-Next-Vicuna-13B, and Yi-VL-34B have prediction distributions that deviate from the actual results by no more than 15\%. Additionally, these four models consistently rank in the top five in terms of performance.
\\
\textbf{Results of Summarization} \quad
In proprietary models, the scores for this task are generally higher, with GPT4V achieving the highest score of 82.40. However, among open-source models, many still have relatively low scores. For example, MiniCPM-Llama3-V-2.5 only score 17.20, with only Phi-3-Vision, Cogvlm2-Llama3-Chat-19B, and LLAVA-Next-Yi-34B scoring above 60.00. In Figure \ref{model prediction example}(d), detailed example of LLAVA-1.5V-7B is presented. LLAVA-1.5v-7B, although detailed, provides an inaccurate and less focused response, meriting a score of 2. 
\\
\textbf{Results of Reasoning} \quad 
GPT4V achieve the best score of 59.40, yet still below a satisfactory level. In Figure \ref{model prediction example}(e), we present examples of responses from GPT4V, LLAVA-V1.6-7B, LLAVA-V1.6-13B, and Cogvlm2-Llama3-Chat-19B regarding reasoning tasks. Cogvlm2-Llama3-Chat-19B provided a more detailed response by repeating the conditions from the question and then indicating the correct next step, which may aid in accurate reasoning. LLAVA-V1.6-7B and LLAVA-V1.6-13B both provide incorrect answers to this question.

\section{Further analysis}\label{sec:furthe-analysis}
In this section, we explore the impact of various factors on the FlowCE benchmark.
\subsection{Model parameter volume}
Among all open-source models, having a larger number of parameters does not necessarily lead to better performance. For instance, the 34B parameter models Yi-VL-34B and LLAVA-Next-Yi-34B scored only 26.24 and 44.33, respectively, while Phi-3-Vision, with only 4.2B parameters, achieved the best score among the open-source models. In Table \ref{table:model_A_scores}, we compare the average performance across three parameter scales. Although there may be a trend of improvement with increasing model parameters, this is not a definitive conclusion.
\subsection{Model data volume}
In Table \ref{table:model-training-dataset}, we provide detailed information on the specific pre-training and fine-tuning data volumes for each model, and further analyze how the data sources impact the performance of model on FlowCE. Despite ShareGPT4V-13B utilizing a larger dataset, its performance still lags behind LLAVA-v1.5-13B, demonstrating that the quality of the dataset is paramount. Additionally, the selection and diversity of specific datasets play a crucial role. For instance, Phi-3-Vision leverages a 0.5T image-text paired dataset that includes FLD-5B, OCR-generated synthetic data, chart comprehension datasets, and plain text data \cite{xiao2024florence, laurenccon2024obelics}. These high-quality and diverse data sources have enabled Phi-3-Vision to achieve the highest score of 35.3 in the information extraction task on FlowCE, and furthermore, it ranks first in the overall score among open-source models.


\begin{table}[h!]
\centering
\small
\renewcommand{\arraystretch}{1.2}
\begin{tabular}{cc}
\toprule
\textbf{Model Parameter} & \textbf{Score} \\
\midrule
3.4B\textasciitilde7B & 32.47 \\
8B\textasciitilde13B & 31.40 \\
13B\textasciitilde & \textbf{35.21} \\
\bottomrule
\end{tabular}
\caption{Average Scores on FlowCE for Different Parameter Levels}
\label{table:model_A_scores}
\end{table}

\subsection{Consensus between Humans and Evaluators}
In this section, we employ manual scoring evaluations for MLLMs' responses in Reasoning, Localization Recognition, and Summarization. The aim is to investigate whether the standards set by FlowCE and the use of GPT4 as an evaluator align closely with human assessment results. We engage five human evaluators to assess the model GPT4o, which emerges as the top-performing model overall. Additionally, we select the open-source model LLAVA-V1.6-13B for manual evaluation. The criteria and detailed results of the manual assessment can be found in the Appendix \ref{appendix:Manual Evaluation Protocol}. In Table \ref{table:model_A_scores}, we present the Pearson correlation coefficients between human ratings and GPT4 scores under our answer setting. The results demonstrate a high degree of consistency between human evaluation and our assessment methodology, indicating that our FlowCE evaluation results can be regarded as effective assessments.
\begin{table}[h!]
    \centering
    \small
    \begin{tabular}{cccc}
    \toprule
     & \textbf{RS} & \textbf{LR} & \textbf{SM}\\
    \midrule
    \textbf{Correlation} & 0.97 & 0.97 & 0.91 \\
    \bottomrule
    \end{tabular}
    \caption{The Pearson correlation coefficient between human ratings and GPT4 scores for various tasks.}
    \label{table:model_A_scores}
\end{table}

\section{Conclusion}
To evaluate the comprehension ability of MLLMs on flowcharts, we propose the first multi-dimensional evaluation method: FlowCE. FlowCE sets up five major categories of tasks, including reasoning, information extraction, localization recognition, logical verification, and summarization, aiming to thoroughly quantify the understanding capability and performance of MLLMs on flowcharts. The FlowCE framework not only provides an effective means to evaluate the comprehension ability of MLLMs on flowcharts, but also offers guidance for model optimization and improvement, thereby promoting the development of MLLMs.

\section*{Limitations}
This work has two limitations. Firstly, it establishes the FlowCE benchmark based on flowcharts derived from a diverse set of 500 real-world images. While it poses challenges for existing closed-source and open-source models, continuous expansion of both the dataset size and the number of questions is necessary going forward. Secondly, FlowCE relies entirely on manual annotation for data generation. However, as the dataset grows, dependence on manual annotation introduces inherent limitations, making it difficult to completely eliminate errors from the data.

\bibliography{custom}

\appendix

\section{Image collection and manual annotation} \label{sec:appendix}
To obtain flowchart images, we first conducted image searches using the keyword "flowchart" on search engine(Baidu Search), and then saved them. However, we encountered issues such as duplicates, low resolution, incomplete images, and other unrelated photos. Therefore, we proceeded to manually select images, resulting in the creation of a real-world dataset.

\begin{figure}[t]
\centering
\includegraphics[width=2.5in]{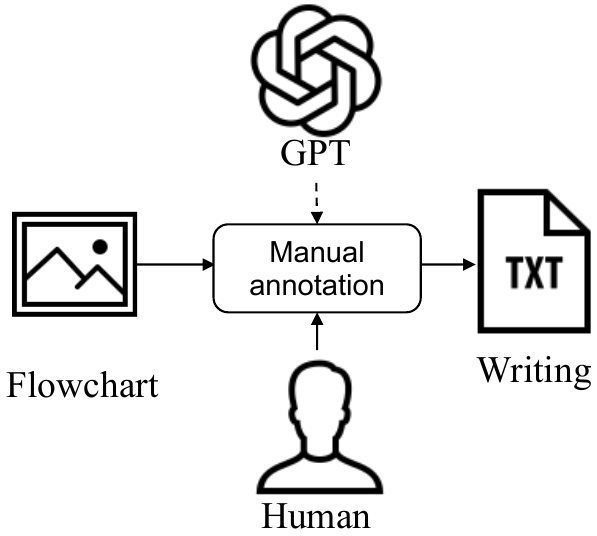}
    \caption{Manual annotation process, with optional assistance from GPT-like models for diversity construction.}
\label{fig:ap1}
\end{figure}

To ensure the construction of question-answer pairs in open scenarios, we use manual annotation for each flowchart. Additionally, to ensure the diversity of the question-answer pairs, we employ a powerful GPT-like model to assist with the generation. The annotation process is illustrated in Figure \ref{fig:ap1}. Humans can choose to use GPT to generate basic diverse question-answer pairs, which are then modified as needed.

\section{Large Language Model Judge} \label{appendix:Large Language Model Judge}
\begin{figure*}[t]
    \centering
    \includegraphics[width=1.00\textwidth]{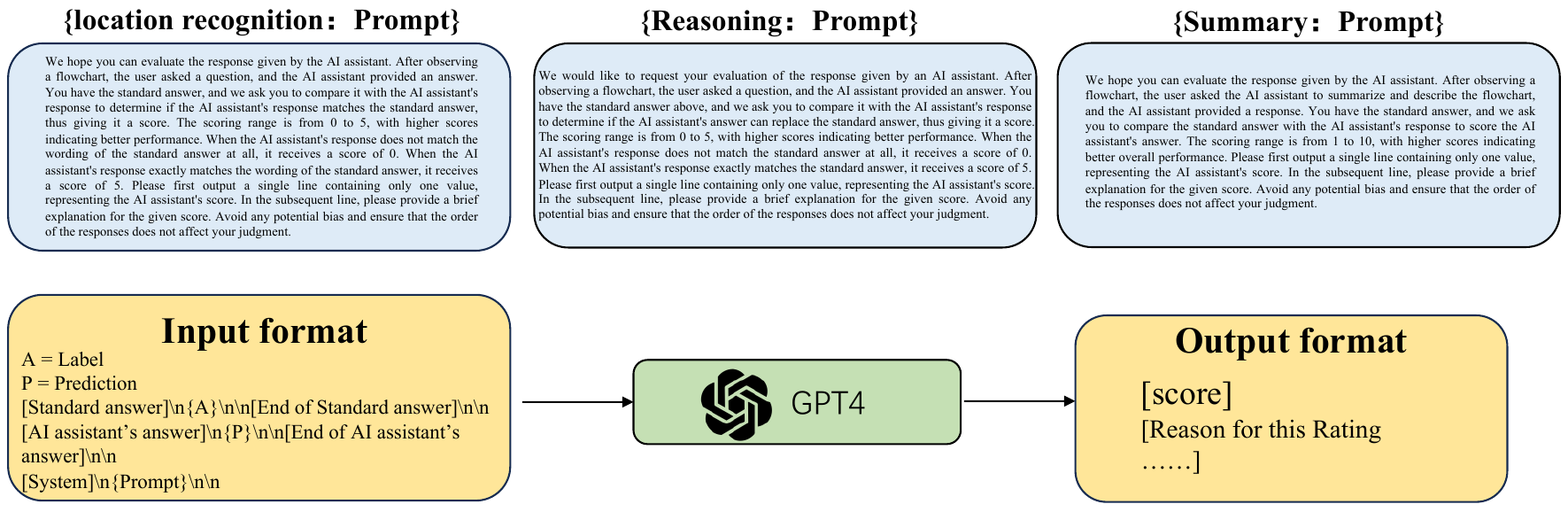} 
    \caption{
Using GPT-4 as an automated evaluator.
}
    \label{eval}
\end{figure*}

We use GPT-4 as an automated evaluator to score tasks in three categories: localization recognition, reasoning, and summarization. The scoring methodology is illustrated in Figure \ref{eval}, where we set a score range of 0-5 for each question in the reasoning and localization recognition tasks, and a score range of 1-10 for the summarization tasks.

\begin{figure*}[t]
    \centering
    \includegraphics[width=0.70\textwidth]{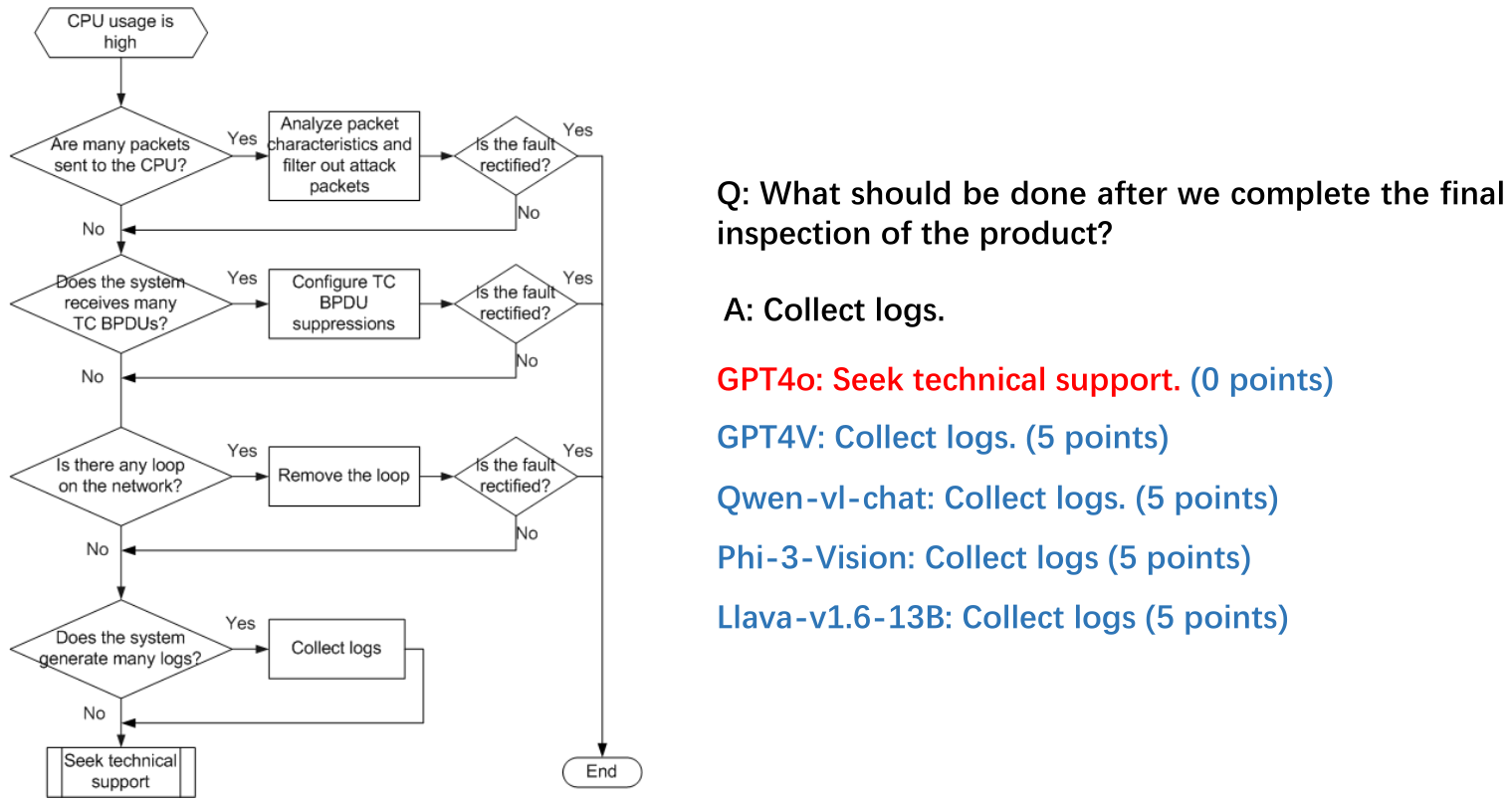} 
    \caption{
Model output and received scores.
}
    \label{1c}
\end{figure*}

\begin{figure*}[t]
    \centering
    \includegraphics[width=0.70\textwidth]{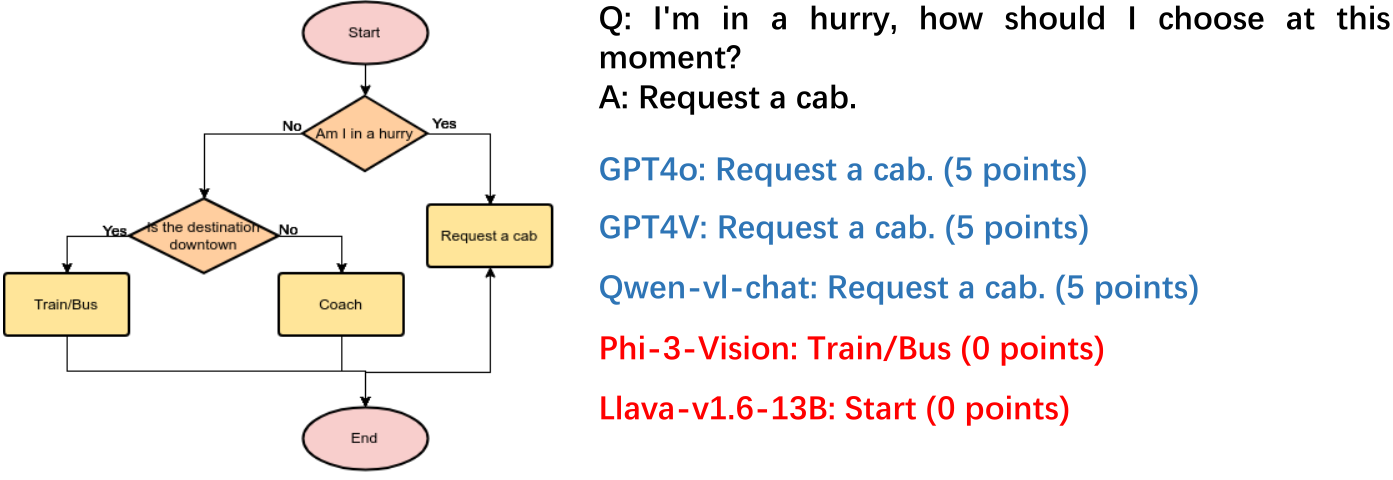} 
    \caption{
Model output and received scores.
}
    \label{2c}
\end{figure*}

\begin{figure*}[t]
    \centering
    \includegraphics[width=0.70\textwidth]{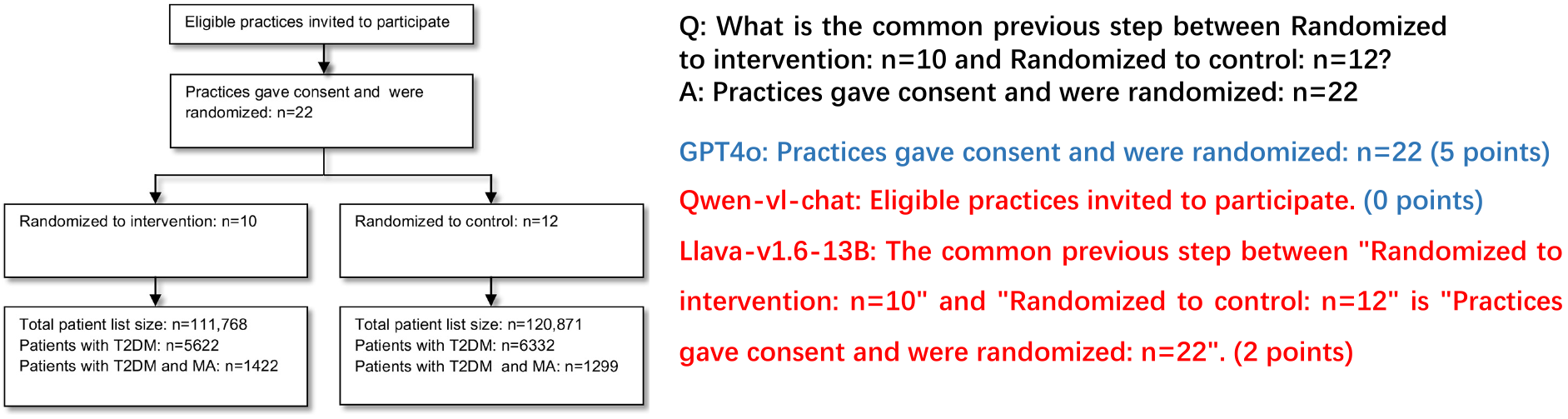} 
    \caption{
Model output and received scores.
}
    \label{3c}
\end{figure*}

\begin{figure*}[t]
    \centering
    \includegraphics[width=0.70\textwidth]{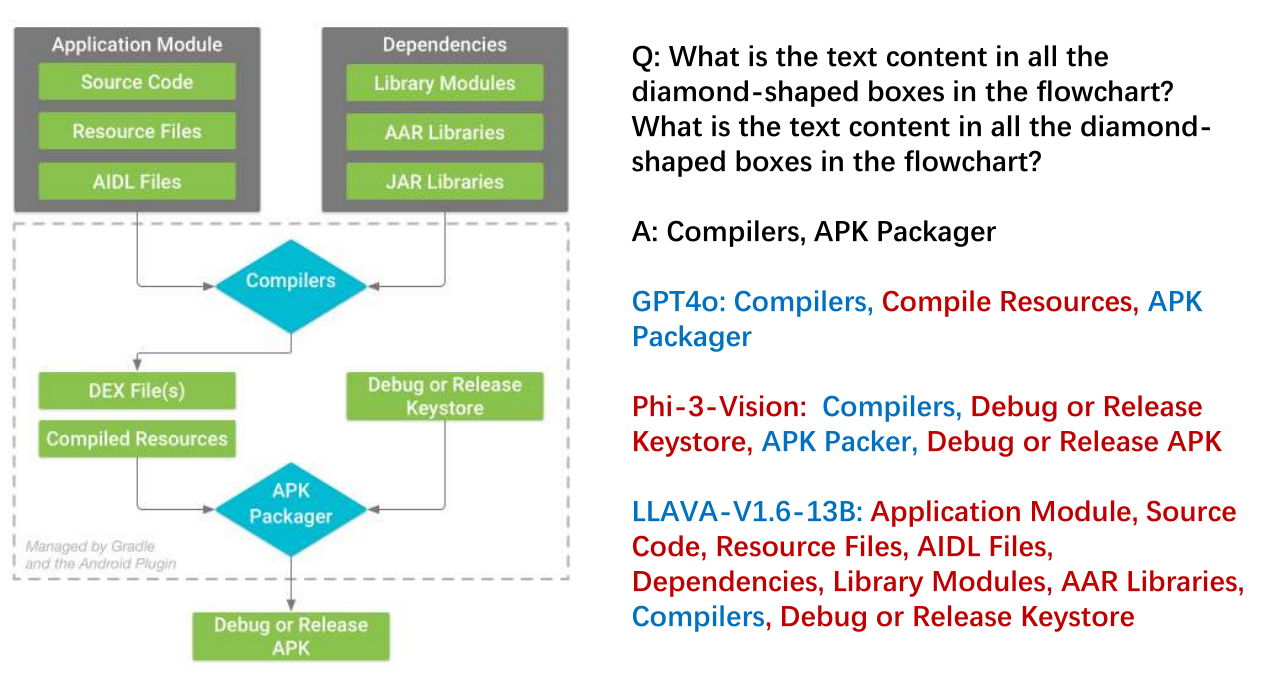} 
    \caption{
Model output and received scores.
}
    \label{4c}
\end{figure*}

\begin{figure*}[t]
    \centering
    \includegraphics[width=0.70\textwidth]{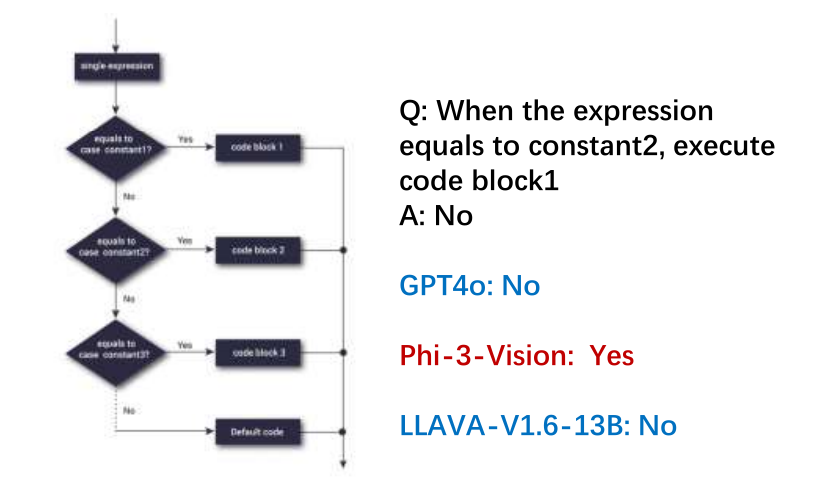} 
    \caption{
Model output and received scores.
}
    \label{5c}
\end{figure*}

\begin{figure*}[t]
    \centering
    \includegraphics[width=0.70\textwidth]{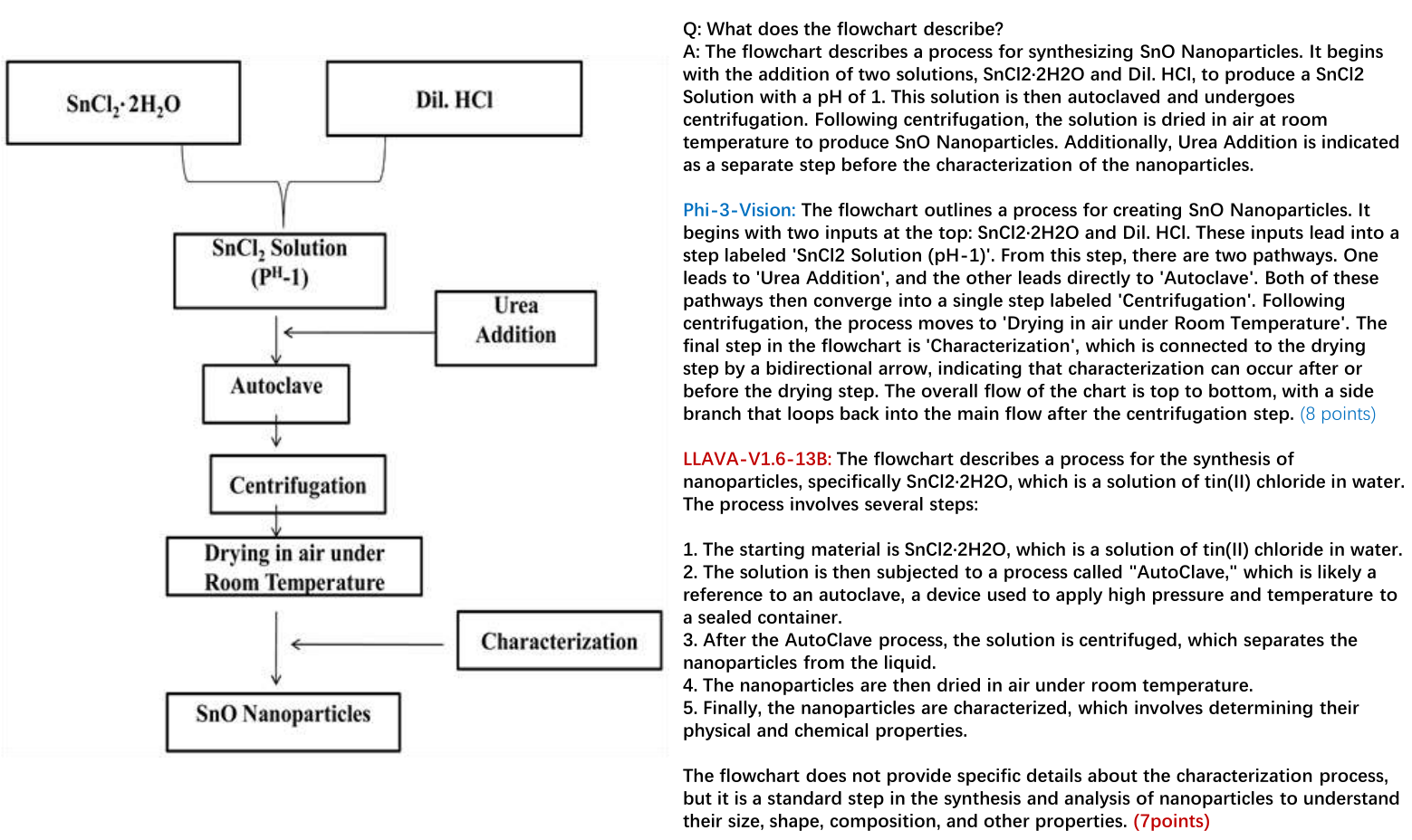} 
    \caption{
Model output and received scores.
}
    \label{6c}
\end{figure*}

\section{Additional examples} \label{appendix:additional example}
We provide more examples in Figures \ref{1c}, \ref{2c}, \ref{3c}, \ref{4c}, \ref{5c} and \ref{6c}, including the model's responses and the evaluation scores given by GPT-4. 

\section{Manual Evaluation Protocol} \label{appendix:Manual Evaluation Protocol}
We provide both the standard answers from FlowCE and the model's output answers. In Table \ref{tab:human-gpt}, we present the detailed scores from the human evaluation.
\\
\textbf{RS and LR Tasks:}
\\
5 points: Answer is entirely correct with no factual errors.
\\
4 points: Answer is mostly correct with minor factual errors that do not affect the main content.
\\
3 points: Answer contains some factual errors but is still useful overall.
\\
2 points: Answer has significant factual errors, with some content being incorrect.
\\
1 point: Answer is severely flawed, with most content being incorrect.
\\
0 points: Completely incorrect.
\\
\textbf{SM Task:} 
\\
10 points: The answer is completely correct without any factual errors.
\\
8 points: The answer is basically correct with only minor factual errors that do not affect the main content.
\\
6 points: The answer contains some factual errors but is still useful on the whole.
\\
4 points: The answer has significant factual errors with incorrect parts of the content.
\\
2 points: The answer is seriously incorrect with most of the content incorrect.
\\
1 point: The answer is completely incorrect.

\begin{table}[h!]
    \footnotesize
    \centering
    \begin{tabular}{cccccccc}
        \toprule
        Model & \multicolumn{3}{c}{GPT4-score} & \multicolumn{3}{c}{Human-score} \\
        \cmidrule(lr){2-4} \cmidrule(lr){5-7}
         & RS & LR & SM & RS & LR & SM \\
        \midrule
        GPT4o & 57.6 & 44.8 & 79.9 & 62.6 & 58.4 & 75.9 \\
        LLAVA & 37.4 & 27.8 & 50.7 & 45.8 & 36.8 & 47.6 \\
        \bottomrule
    \end{tabular}
    \caption{Performance Comparison between GPT4o and LLAVA.}
    \label{tab:human-gpt}
\end{table}

\section{Detailed Comparison of Logical Verification Task} \label{appendix:predicted_vs_actual}
In this section, we present a detailed comparison of the predicted results for a subset of models: GPT4o, Phi-3-Vision, LLaVA-Next-Vicuna-13B, Qwen-Chat-VL, LLaVA-Next-Vicuna-7B, LLaVA-V1.5-13B, Cogvlm2-Llama3-Chat-19B, and Cogvlm-Chat. 
\begin{figure*}[h]
    \centering
    \includegraphics[width=1.00\textwidth]{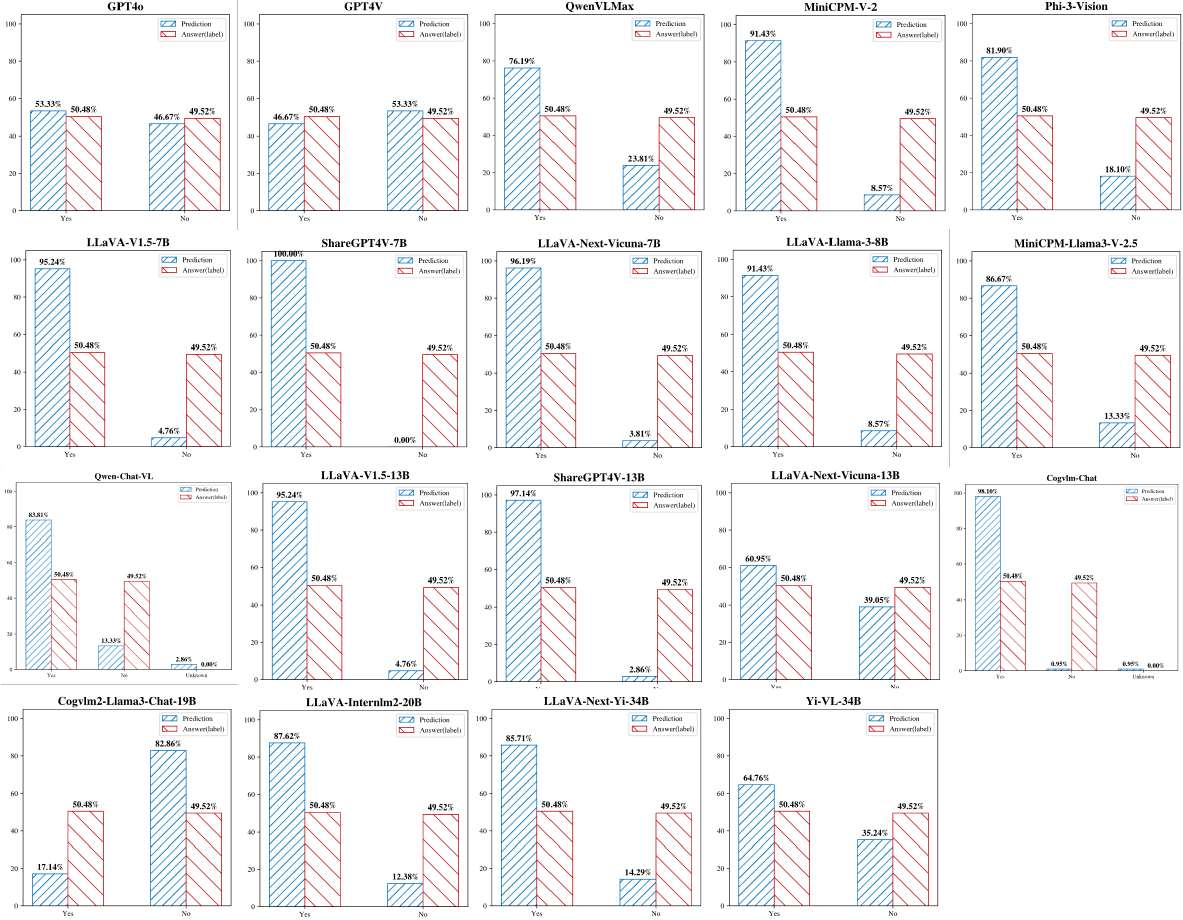} 
    \caption{
The prediction distributions of different models on the logical verification task.
}
    \label{bar}
\end{figure*}
Each subplot in Figure \ref{bar} compares the predicted results (in blue) with the actual answer labels (in red) for each model. The score below each subplot indicates the overall performance of the model based on its accuracy in predicting the correct category.

GPT4o stands out with the highest accuracy, achieving a score of 83.81, indicating robust performance in aligning predictions with actual labels. Phi-3-Vision, while scoring 60.95, demonstrates a noticeable discrepancy in the \textit{"No"} category with lower prediction accuracy. LLaVA-Next-Vicuna-13B, with a score of 62.86, shows moderate alignment but also exhibits substantial errors in the \textit{"No"} category. Qwen-Chat-VL and Cogvlm-Chat, both scoring 50.48, indicate significant prediction errors and lower overall accuracy, particularly evident in the \textit{"No"} and \textit{"Unknown"} categories. LLaVA-Next-Vicuna-7B and LLaVA-V1.5-13B, scoring 52.38 and 53.55 respectively, also reflect moderate performance but with specific inaccuracies in the \textit{"No"} category. Cogvlm2-Llama3-Chat-19B, with a score of 57.14, shows better performance than some other models but still falls short in accurately predicting the \textit{"No"} responses. These results suggest that while certain models like GPT4o exhibit strong performance, others require significant improvements in understanding and predicting both \textit{"Yes"} and \textit{"No"} categories accurately. The varying scores underscore the necessity for further refinement in training methodologies and model architectures to enhance predictive accuracy across all categories.

The phenomenon where some models exhibit a near 100\% probability in answering \textit{"Yes"} can be attributed to several factors:

\begin{figure*}[h]
    \centering
    \includegraphics[width=1.00\textwidth]{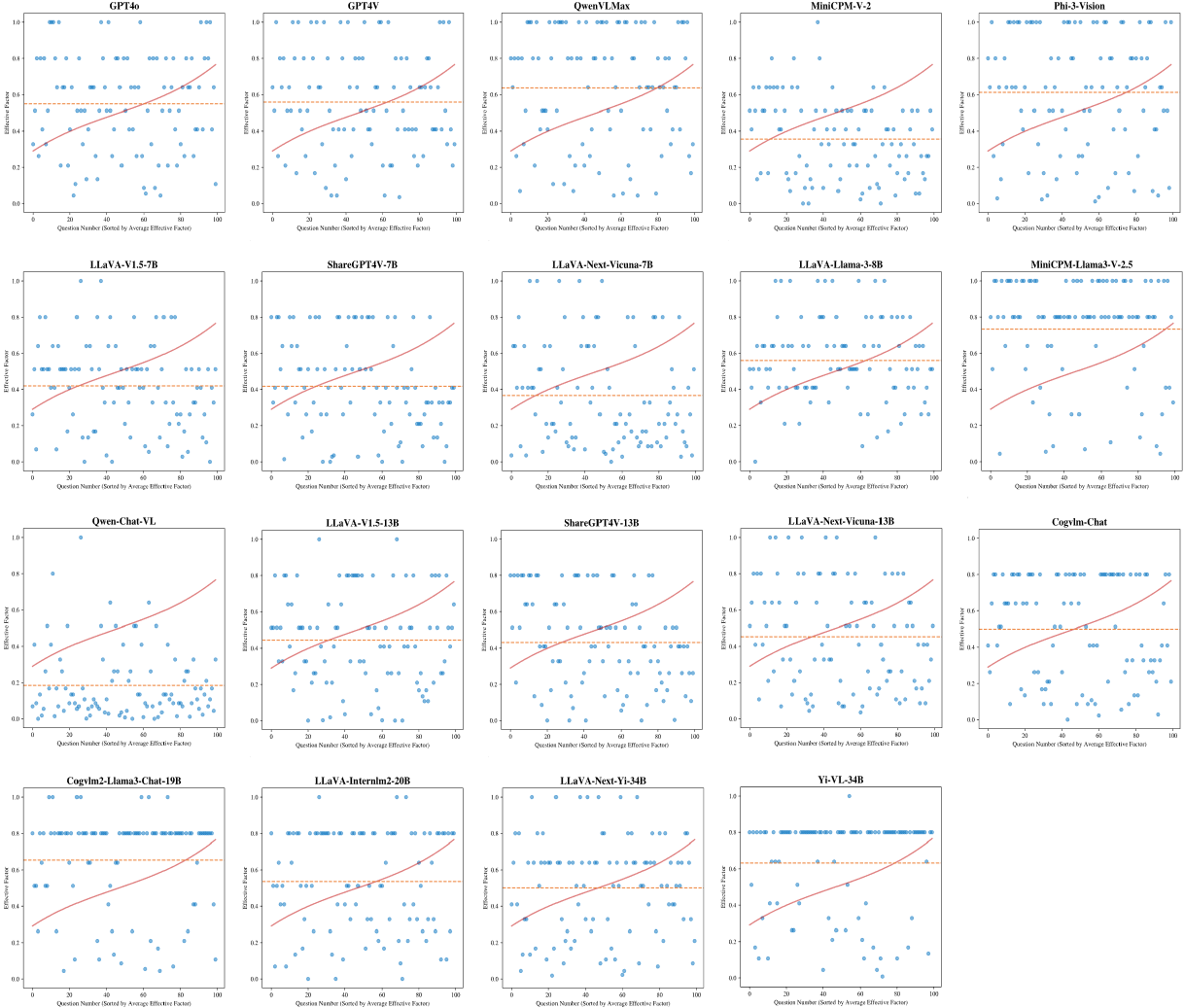} 
    \caption{
Comparison of MLLMs Performance on Information Extraction tasks based on effective factor distribution. The red line represents the smoothed ascending order of the average effective factor across all models for each specific question. The orange line indicates the average effective factor for each model across all question.
}
    \label{scatter}
\end{figure*}

\begin{itemize}
    \item \textbf{Training Data Bias}: The training datasets may have an imbalance where affirmative answers (\textit{"Yes"}) are disproportionately represented compared to negative ones (\textit{"No"}). This bias in the training data can lead the models to favor \textit{"Yes"} responses, as they learn to associate the affirmative answer with higher probabilities during the training process.
    
    \item \textbf{Model Overfitting}: Certain models might be overfitted to specific patterns in the training data, especially if those patterns predominantly involve affirmative responses. Overfitting can cause the model to generalize poorly to new, unseen data, resulting in a high likelihood of predicting \textit{"Yes"} regardless of the actual context or question.
    
    \item \textbf{Algorithmic Tuning}: The hyperparameters and algorithmic settings of some models might be tuned in a way that inadvertently biases the model towards affirmative responses. This could include settings related to decision thresholds, loss functions, or other optimization parameters that skew the model's predictions towards \textit{"Yes"}.
    
    \item \textbf{Lack of Contextual Understanding}: Some models may lack the nuanced understanding required to accurately discern between \textit{"Yes"} and \textit{"No"} in complex scenarios. This deficiency can lead them to default to a \textit{"Yes}" answer, especially if they are not effectively capturing and processing the context of the queries.
    
    \item \textbf{Evaluation Metrics}: The evaluation metrics used during the training and validation phases might inadvertently prioritize accuracy in affirmative answers due to the distribution of the training data. This focus on affirmative accuracy can lead the models to perform better on \textit{"Yes"} predictions, inflating the probability of such responses in practical applications.
\end{itemize}

These factors collectively contribute to the observed high probability of \textit{"Yes"} responses in some models, highlighting the need for balanced training data, careful tuning, and improved contextual understanding in model development.

\section{Detailed Comparison of Informaton Extraction Task} \label{appendix:Informaton-Extraction-task}

As shown in Figure \ref{scatter}, there are notable differences in the distribution of effective factor values among various models. Phi-3-Vision and GPT-4 have a broad distribution of effective factor values, with more data points in the higher effective factor region (above 0.6). Other models have a more scattered distribution of effective factor values, with most concentrated in the low effective factor region, particularly Qwen-Chat-VL, where the majority of data points are below 0.2. Despite the Cogvlm family having relatively high effective factors and fewer incorrect answers, the overall number of labels is also small. After subtracting the incorrect labels, the number of correct labels is minimal, resulting in a very low total score.

\end{document}